\documentclass[11pt,a4paper]{article}
\usepackage[hyperref]{acl2021}
\usepackage{times}
\usepackage{latexsym}
\pdfoutput=1

\usepackage{booktabs} 
\usepackage{multirow}
\usepackage{tikz-dependency}
\usepackage{amssymb}
\usepackage{mathtools}
\usepackage[linesnumbered,ruled,vlined,noend]{algorithm2e}

\usepackage{microtype}

\aclfinalcopy 

\newcommand{\compacturl}[1]{{\fontsize{10}{8.5}\selectfont\textls*[-70]{\url{#1}}}}
\newcommand{\compacturll}[1]{{\fontsize{4}{3}\selectfont\textls*[-100]{\url{#1}}}}

\title{{COMBO}: a new module for {EUD} parsing}

\author{Mateusz Klimaszewski$^{1,2}$ Alina Wr\'{o}blewska$^2$  \\
$^1$Warsaw University of Technology \\ 
$^2$Institute of Computer Science, Polish Academy of Sciences \\
{\tt m.klimaszewski@ii.pw.edu.pl} \\ {\tt alina@ipipan.waw.pl}
}

\date{}

\begin{document}
\maketitle
\begin{abstract}
We introduce the~COMBO-based approach for EUD parsing and its implementation, which took part in the IWPT 2021 EUD shared task. The goal of this task is to parse raw texts in 17 languages into Enhanced Universal Dependencies (EUD). The proposed approach uses COMBO to predict UD trees and EUD graphs. These structures are then merged into the~final EUD graphs. Some EUD edge labels are extended with case information using a~single language-independent expansion rule. In the official evaluation, the~solution ranked fourth, achieving an~average ELAS of 83.79\%. The~source code is available at \url{https://gitlab.clarin-pl.eu/syntactic-tools/combo}.
\end{abstract}

\section{Introduction}
Data-driven dependency parsers achieve high parsing performance for languages representing different language families. The~state-of-the-art dependency parsers are trained with supervised learning methods on large correctly annotated treebanks, e.g. from Universal Dependencies \cite[UD,][]{nivre-etal-2020-universal}. UD is an~international initiative aimed at developing a~cross-linguistically consistent annotation schema and at building a~large multilingual collection of dependency treebanks annotated according to this schema.
A~relatively small subset of UD treebanks is annotated with higher-order syntactic-semantic representations that encode various linguistic phenomena and are called Enhanced Universal Dependencies (EUD).

Dependency treebanks, especially the~uniformly annotated UD treebanks, are used for multilingual system development, e.g. within multiple shared tasks on dependency parsing \cite{buchholz-marsi-2006-conll,nivre-etal-2007-conll,seddah-etal-2013-overview,seddah-etal-2014-introducing,zeman-etal-2017-conll,zeman-etal-2018-conll}. In particular, the~IWPT 2020 shared task on Parsing into Enhanced Universal Dependencies \cite{bouma-etal-2020-overview} is worth mentioning, because it is the predecessor of the~current IWPT 2021 shared task \cite{bouma-etal-2021-overview}. All shared tasks contributed to rapid advancement of language parsing technology, inter alia, the formulation of groundbreaking parsing algorithms and their publicly available implementations \cite[e.g.][]{nivre-etal-2006-labeled,mcdonald-etal-2006-multilingual,straka-strakova-2017-tokenizing,dozat-etal-2017-stanfords,rybak-wroblewska-2018-semi,he-choi-2020-adaptation}.

Dependency parsing is an~important issue in various sophisticated downstream tasks, including but not limited to sentiment analysis \cite{sun-etal-2019-aspect}, relation extraction \cite{zhang-etal-2018-graph,vashishth-etal-2018-reside,guo-etal-2019-attention}, semantic role labelling \cite{wang-etal-2019-best}, or question answering \cite{AAAI1817406}. On the other hand, even if EUD parsing aims at predicting semantically informed structures, which seem to be appropriate in advanced NLP tasks, it is not yet used in solving these tasks. An~obstacle can be the~availability of the~state-of-the-art EUD parsers, e.g. two top systems at the~IWPT 2020 EUD shared task \cite[i.e.][]{kanerva-etal-2020-turku,heinecke-2020-hybrid} are not publicly available and therefore difficult to integrate into NLU systems
without having to implement them from scratch. Meeting the~potential expectations of NLU system architects, the~source code of COMBO with the~new EUD parsing module and the pre-trained models developed as part of our solution submitted to this shared task are publicly available.

The~proposed solution to EUD parsing is based on (1) Stanza tokeniser \cite{qi-etal-2020-stanza}, (2) COMBO \cite{combo}, a~data-driven language-independent system for morphosyntactic prediction, i.e. part-of-speech tagging, morphological analysis, lemmatisation, dependency parsing, and EUD parsing (see Section \ref{sec:eud_predictor}), (3) an~algorithm that merges predicted labelled dependency arcs and predicted EUD arcs, and builds the~final EUD graphs (see Section \ref{sec:merge_algo}), and (4) two linguistically motivated language-independent rules that improve the~final EUD graphs (see Section \ref{sec:postprocessing}). The~first expansion rule adds case information sublabels to EUD modifiers, and the~second one amends enhanced arcs coming into the~function words. These two rules are integrated into the~proposed EUD parsing system.

In the~official evaluation, our EUD parser ranked 4th, obtaining an~average ELAS of 83.79\% and EULAS of 85.20\%.\footnote{\url{https://universaldependencies.org/iwpt21/results_official_coarse.html}} 
It is worth emphasising that COMBO predicts labelled dependency trees with an~average LAS of 88.91\%, only being slightly outperformed by the~ROBERTNLP system.

\section{Shared task description}
The~IWPT 2021 EUD shared task consists in evaluating systems for parsing raw texts into Enhanced Universal Dependencies. The~systems are trained and evaluated on data supplied by the~organisers.

\paragraph{Data} The~shared task dataset includes treebanks for 17 languages from 4 language families. The~largest group in this collection is constituted by Indo-European languages, i.e. Bulgarian, Czech, Polish, Russian, Slovak, Ukrainian (Slavic), Dutch, English, Swedish (Germanic), French, Italian (Romance), and Latvian, Lithuanian (Baltic). There are also representatives of the~Uralic (Finnic) languages, i.e. Estonian and Finnish, the~Afro-Asiatic (Semitic) languages -- Arabic, and the~Southern Dravidian languages -- Tamil. The~datasets vary in size and type of enhancements.

\begin{figure}[h!]
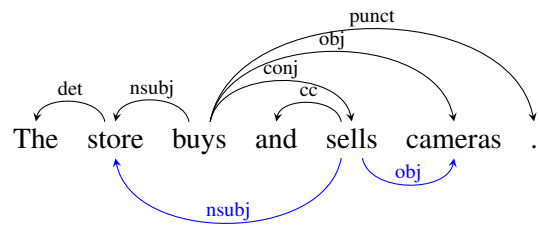

\centering
\begin{dependency}[theme = simple, label style={font=\normalsize}]
   \begin{deptext}[column sep=0.2cm]
      The \& store \& buys \& and \& sells \& cameras \& . \\
   \end{deptext}
   \depedge{2}{1}{det}
   \depedge
   {3}{2}{nsubj}
   \depedge{3}{5}{conj}
   \depedge{3}{7}{punct}
   \depedge{5}{4}{cc}
   \depedge{3}{6}{obj}
   \depedge[edge style=blue, edge below]{5}{2}{\textcolor{blue}{nsubj}}
   \depedge[edge style=blue, edge below
   ]{5}{6}{\textcolor{blue}{obj}}
\end{dependency}
\caption{The~EUD graph with the~conjoined predicate; the~conjoined verbs (\textit{buys} and \textit{sells}) share the~subject (\textit{the store}) and the~object (\textit{cameras}), and the~propagated relations are indicated with the bottom blue enhanced edges.}
\label{fig:conjuncts}
\end{figure}

\begin{figure}[h!]
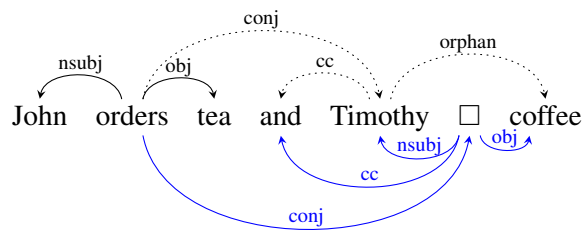

\centering
\begin{dependency}[theme = simple, label style={font=\normalsize}]
   \begin{deptext}[column sep=0.2cm]
      John \& orders \& tea \& and \& Timothy \& $\square$  \& coffee \\
   \end{deptext}
   \depedge{2}{1}{nsubj}
   \depedge{2}{3}{obj}
   \depedge[edge style=dotted]{2}{5}{conj}
   \depedge[
   dotted]{5}{7}{orphan}
   \depedge[edge style=dotted]{5}{4}{cc}
   \depedge[edge style=blue, edge below,
   ]{6}{5}{\textcolor{blue}{nsubj}}
   \depedge[edge style=blue, edge below, edge end x offset=-6pt]{6}{7}{\textcolor{blue}{obj}}
   \depedge[edge style=blue, edge below,
   ]{2}{6}{\textcolor{blue}{conj}}
   \depedge[edge style=blue, edge below,
   ]{6}{4}{\textcolor{blue}{cc}}
\end{dependency}
\caption{The EUD graph with an~empty node $\square$ and the bottom blue enhanced edges. The~tree edges removed from the~EUD graph are dotted.}
\label{fig:emptynode}
\end{figure}

\begin{figure}[h!]
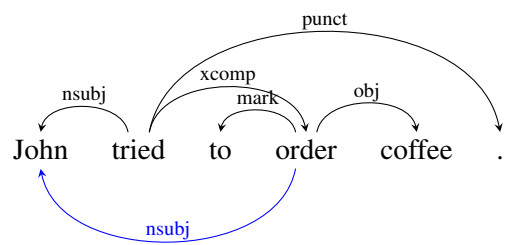

\centering
\begin{dependency}[theme = simple, label style={font=\normalsize}]
   \begin{deptext}[column sep=0.4cm]
      John \& tried \& to \& order \& coffee \& . \\
   \end{deptext}
   \depedge{2}{1}{nsubj}
   \depedge{2}{4}{xcomp}
   \depedge{4}{3}{mark}
   \depedge{4}{5}{obj}
   \depedge{2}{6}{punct}
   \depedge[edge style=blue, edge below 
   ]{4}{1}{\textcolor{blue}{nsubj}}
\end{dependency}
\caption{The EUD graph with the~bottom blue enhanced edge encoding subject control with the~control predicate \textit{try}.}
\label{fig:controll}
\end{figure}

\begin{figure}[h!]
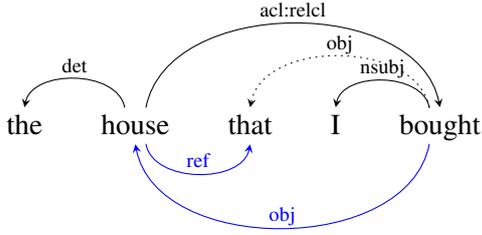

\centering
\begin{dependency}[theme = simple, label style={font=\normalsize}]
   \begin{deptext}[column sep=0.6cm]
      the \& house \& that \& I \& bought \\
   \end{deptext}
   \depedge{2}{1}{det}
   \depedge{2}{5}{acl:relcl}
   \depedge{5}{4}{nsubj}
   \depedge[edge style=dotted]{5}{3}{obj}
   \depedge[edge style=blue, edge below
   ]{2}{3}{\textcolor{blue}{ref}}
   \depedge[edge style=blue, edge below]{5}{2}{\textcolor{blue}{obj}}
\end{dependency}
\caption{The EUD graph representing a~relative clause modifying the noun \textit{house}. The~enhanced edges are marked with the~bottom blue arcs and the~tree edge removed from the~EUD graph is dotted.}
\label{fig:relative_clause}
\end{figure}

\paragraph{Enhancement types}
\label{sec:enhancemets}
Various linguistic phenomena are encoded in EUD graphs:
\begin{itemize}
\item propagation of conjuncts in coordination constructions (see Figure \ref{fig:conjuncts}),
\item null nodes encoding elided predicates in coordination constructions (see Figure \ref{fig:emptynode}),
\item additional subject relations in control and raising constructions (see Figure \ref{fig:controll}),
\item coreference relations in relative clause constructions (see Figure \ref{fig:relative_clause}),
\item detailed case information sublabels of the~modifiers (see Figure \ref{fig:case}).
\end{itemize}

\section{System overview}
The~EUD parsing system is built of the~following components: a~data encoder boosted with a~contextual language model (see Section \ref{sec:dataencoder}), morphosyntactic predictors (see Section \ref{sec:predictor_architecture}), an~EUD predictor (see Section \ref{sec:eud_predictor}), an~algorithm merging predicted labelled dependency arcs and enhanced dependency arcs (see Section \ref{sec:merge_algo}), and a~post-processing module (see Section \ref{sec:postprocessing}).

\subsection{Data encoder}
\label{sec:dataencoder}

The~encoder vectorises the~tokenised input data. The~input tokens are first represented as a~concatenation of a~character-based word embedding estimated during system training with a~dilated convolutional neural network \cite{dcnn:2015}, and a~BERT-based embedding estimated as follows. 

\noindent
BERT-based language models \cite[LM,][]{devlin-etal-2019-bert,conneau-etal-2020-unsupervised} are not fine-tuned during system training. Instead, we apply the scalar mix technique based on \citet{peters-etal-2018-deep} to produce an~embedding ($h$) for a~
word $i$ as a~weighted sum of embeddings from all layers:
\begin{equation}
    h_i = \gamma \sum_{j=1}^L s_j h_{ij}
\end{equation}
Parameters $\gamma$ and $s_j$ are learnable weights, additionally $s_j$ are softmax-normalised. $L$ is the number of transformer layers. At the point of using LM, the~data is already tokenised. If LM intra-tokeniser splits a~word into multiple subwords, the~embeddings $h$ are estimated for these subwords and averaged. The~vectors of words or averaged vectors of subwords are finally transformed with one fully connected (FC) layer.

The~encoder with two BiLSTM layers \cite{lstm:1997,bilstm:2005} 
transforms the~concatenations of the~character-based word embeddings and the~transformed BERT-based embeddings into token vectors. The~BiLSTM-transformed token embeddings are used as input to morphosyntactic predictors and the~EUD parsing module.

\subsection{Morphosyntactic predictors}
\label{sec:predictor_architecture}
The~proposed approach is based on various morphosyntactic predictions. Part-of-speech tags, morphological features, and lemmata are used in the~post-processing step to extract case information expanding enhanced sublabels of modifiers (see Section \ref{sec:postprocessing}). The~merge algorithm (see Section \ref{sec:merge_algo}), in turn, combines labelled dependency arcs with enhanced dependency arcs predicted by EUD parsing module. 

\subsection{EUD predictor}
\label{sec:eud_predictor}
The~EUD parsing module consists of an~enhanced arc classifier and an~enhanced label classifier. The~
arc classifier utilises two single FC layers that transform encoded token vectors
into head and dependent embeddings. These embeddings are used to calculate an~adjacency matrix ($A$) of an~enhanced graph. $A$ is a~$n \times n$ matrix, where $n$ is the~number of tokens in a~sentence (plus the~\textsc{root} node). The~matrix element $A_{ij}$ corresponds to the~dot product of the~$i$-th dependent embedding and the~$j$-th head embedding. The~dot product indicates the~certainty of the~edge between two tokens. The sigmoid function, applied to each element of $A$, allows the network to predict many heads for a given dependent, i.e. EUD graphs are built.

\noindent
The~enhanced label classifier also applies two fully connected layers to estimate head ($e_i$) and dependent ($e_j$) embeddings (they differ from embeddings estimated in the~enhanced arc prediction). Enhanced dependency labels are predicted by a~fully connected layer with the~softmax activation function which is given the~dependent embedding concatenated with the~head embedding. 
\begin{equation}
    e_{head} = \mathit{FC}(e_i)
\end{equation}
\begin{equation}
    e_{dep} = \mathit{FC}(e_j)
\end{equation}
\begin{equation}
    \mathit{label} = \mathit{argmax}(\mathit{FC}(e_{head}, e_{dep}))
\end{equation}
The loss function is only propagated for those pairs ($i, j$) that belong to ground truth (i.e. arcs existing in the enhanced dependency graph).

\subsection{Merge algorithm}
\label{sec:merge_algo}

The~predicted enhanced graphs could be used without further processing. However, their quality could definitely be improved if they exploited information from the~predicted dependency trees.
Enhanced dependency graphs appear to be heavily tree-based (see the~example EUD graphs in Section \ref{sec:enhancemets}).
The~EUD graphs include some additional edges, empty nodes, and extended labels of modifiers (and conjuncts in some languages), or their structure is slightly transformed. We therefore decided to merge the~predicted trees and the~predicted enhanced graphs.

\begin{algorithm}
\SetAlgoLined
\SetKwInOut{Input}{Input}  
\Input{$\:T\coloneqq(V, E_T): \text{tree}$\\
$\:G\coloneqq(V, E_G): \text{graph}$\\
}
\SetKwInOut{Output}{Output}  
\Output{$\:\textit{EUD}: \text{the~final EUD graph}$}
$E_{\textit{EUD}} = \{\}$\;
\For{$e \: \text{in} \: E_T$} {
  \If{$\text{label}(e) \neq \text{acl:relcl}$}{
    $E_{\textit{EUD}} \coloneqq E_{\textit{EUD}} + e $\;
    }
}
\For{$e \: \text{in} \: E_G$} {
  \If{$e \notin E_{\textit{EUD}}$ {\bf and} $\text{has\_no\_cycle}(E_{\textit{EUD}} + e)$}{
    $E_{\textit{EUD}} \coloneqq E_{\textit{EUD}} + e$\;
    }
}
\For{$e \: \text{in} \: E_T$} {
  \If{$\text{label}(e) = \text{acl:relcl}$}{
    $E_{\textit{EUD}} \coloneqq E_{\textit{EUD}} + e $\;
    }
}
$\textit{EUD} \coloneqq (V, E_{\textit{EUD}})$\;
 \caption{The merge algorithm}
 \label{algo2}
\end{algorithm}

The merge algorithm (see Algorithm \ref{algo2}) successively adds the~predicted tree and graph edges to the~set of EUD edges, and then composes the~final EUD graph of these edges. It starts by selecting all tree edges except for edges with the~\textit{acl:relcl} label. The~EUD graphs representing relative clauses contain cycles (see Figure \ref{fig:relative_clause}). Refraining from adding the~\textit{acl:relcl} relations in this step, we attempt to avoid the~cycle problem thereafter. In the~second step, consecutive graph edges are added to the~EUD set as long as they do not form a~cycle or there are no edges with the~same or a different label in the~EUD set (i.e. we eliminate duplicate edges). In the~last step, the~\textit{acl:relcl} relations are added to the~EUD set which is then used to compose a~final EUD graph.

We are aware that UD relations selected in the~first merging step do not contain case information, e.g. the~\textit{obl} relation is transferred to the~EUD set, although this relation should be de facto labelled \textit{obl:because\_of}, \textit{obl:for}, or \textit{obl:outside}. However, our preliminary experiments indicated that the~anticipated enhanced labels often had erroneous case extensions, which could not even come from a~sentence. 
Correcting labels with accidental case extensions would require defining a~large number of relabelling rules that would have to be adapted to a~particular language.  
Extending the~modifier labels rather than correcting them seems to be a~more transparent and simple procedure. We thus define one rule that derives case information from automatically predicted morphological features and lemmata (see Rule 1 in Section \ref{sec:postprocessing}). The rule is utilised in the~post-processing step, which is the~last step of building the~EUD graphs.

\subsection{Post-processing}
\label{sec:postprocessing}
\begin{figure}[h!]
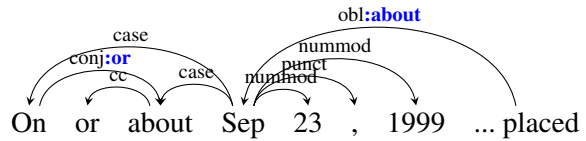

\centering
\begin{dependency}[theme = simple, label style={font=\normalsize}]
   \begin{deptext}[column sep=0.2cm]
      On \& or \& about \& Sep \& 23 \& , \& 1999 \& ... placed \\
   \end{deptext}
   \depedge{4}{1}{case}
   \depedge{4}{3}{case}
   \depedge{3}{2}{cc}
   \depedge{1}{3}{conj\textcolor{blue}{\textbf{:or}}}
   \depedge{4}{5}{nummod}
   \depedge{4}{6}{punct}
   \depedge{4}{7}{nummod}
   \depedge{8}{4}{obl\textcolor{blue}{\textbf{:about}}}
\end{dependency}
\caption{The EUD graph with blue, bolded sublabels representing case information. The~text excerpt comes from the~sentence "On or about September 23, 1999 a request for service was placed by the above referenced counterparty.".}
\label{fig:case}
\end{figure}

We define two rules that improve the~automatically predicted EUD graphs.
\paragraph{Rule 1} The first rule specifies case information of the~following modifiers: \textit{nmod} (nominal modifier), \textit{obl} (oblique nominal), \textit{acl} (clasual modifier of nouns), \textit{advcl} (adverbial clause modifier), and of conjuncts (\textit{conj}). The~case information (lemma) is derived from \textit{case}/\textit{mark} or \textit{cc} dependents of a~modifier or a~conjunct, respectively, and from the~modifier's morphological attribute \textit{Case}. Figure \ref{fig:case} exemplifies extending UD labels with case information.\footnote{This~sentence originates from the~English dev set. As the~case extension of the~\textit{obl} label is derived from the~structure coordinating two prepositions (i.e. \textit{on} and \textit{about}), we wonder about correctness of selecting only \textit{about} as the case extension.}

The~rule is language-independent and UD-based. However, as not all treebanks attribute case information to their modifiers or conjuncts, the rule applies only to predefined languages, e.g. the~conjunct extension is only valid in English, Italian, Dutch, and Swedish.

\paragraph{Rule 2} The~second rule corrects enhanced edges coming into the~function words that are labelled \textit{mark}, \textit{punct}, \textit{root}, \textit{case}, \textit{det}, \textit{cc}, \textit{cop}, \textit{aux} and \textit{ref}. They should not be assigned other dependency relation types in EUD graphs. If a~token \textit{and} is assigned the~\textit{cc} 
grammatical function in a~dependency tree, and thus also in the~corresponding EUD graph (the~first merge step), it cannot be simultaneously a~subject (\textit{nsubj}), for example. If such an~erroneous \textit{nsubj} relation exists, it is removed from the~EUD graph in line with the~second rule. 

\begin{table*}[ht!]
\centering
\begin{tabular}{lll}
\toprule
Language & Model name & Reference\\
\midrule
Arabic & bert-base-arabertv2 & \citet{antoun-etal-2020-arabert} \\
English & bert-base-cased & \citet{devlin-etal-2019-bert} \\
French & camembert-base & \citet{martin-etal-2020-camembert} \\
Finnish & bert-base-finnish-cased-v1 & \citet{DBLP:journals/corr/abs-1912-07076} \\
Polish & herbert-large-cased & \citet{mroczkowski-etal-2021-herbert} \\
\hline
Others & xlm-roberta-large & \citet{conneau-etal-2020-unsupervised} \\
\bottomrule
\end{tabular}
\caption{\label{tab:berts} Language models used in the experiments. Names refer to Transformers library \cite{wolf-etal-2020-transformers}.}
\end{table*}

\section{Experimental setup}
\subsection{Segmentation and preprocessing}
\label{sec:segmentation}
Stanza tokeniser \cite{qi-etal-2020-stanza} is used to split raw text into sentences, split sentences into tokens, and optionally 
to expand multi-words. We train a~new segmentation model for each language on the~training data provided 
in the shared task.\footnote{It is not allowed to use versions of UD other than 2.7 in the IWPT 2021 shared task (see \compacturl{https://universaldependencies.org/iwpt21/task_and_evaluation.html}). As the~publicly available Stanza models are trained on UD 2.5, we have to train new models on UD 2.7.} Whenever there are several UD treebanks for a language, we train the segmentation model on the concatenation of all training datasets available for that language. Multi-word expansion involves only two languages, i.e. Arabic and Tamil, because it does not cause substantial gains in parsing other languages.  

In order to collapse empty nodes, training data are preprocessed with the~official UD script.\footnote{\compacturl{https://github.com/UniversalDependencies/tools/blob/master/enhanced_collapse_empty_nodes.pl}} Dependents of the collapsed empty nodes are assigned new labels, corresponding to the~empty node label and the~dependent label joined with the~special symbol $>$. During prediction, the~collapsed labels are expanded and empty nodes are added at the~end of a~sentence, following \citet{he-choi-2020-adaptation}. This design decision is motivated by the~fact that (1) it is difficult to find a~proper position of elided tokens or phrases, especially in free word order languages, and (2) the~evaluation procedure does not take an~empty node position into account, i.e. appending an empty node at the end of a sentence does not downgrade the score. It is important to note that designing a~heuristic that identifies proper positions of elided elements remains an~open issue, and appending empty nodes at the~end of a~sentence is only a~makeshift solution.

Input data are encoded using BERT-based language models. Depending on the language, either language-specific BERT \cite{devlin-etal-2019-bert} or multilingual XLM-R \cite{conneau-etal-2020-unsupervised} is used (see 
Table \ref{tab:berts}).

\subsection{Morphosyntactic prediction}

COMBO system \cite{combo} is used to predict part-of-speech tags, morphological features, lemmata, and dependency trees.
For the~purpose of this task, we also implement a~new EUD parsing module (see Section \ref{sec:eud_predictor}) and integrate it with COMBO. 
Similarly to segmentation models, we train one COMBO model for a~language on all treebanks provided for this language in the~shared task data using the~default training parameters (see Table \ref{tab:hyper_train}).\footnote{All models are trained and tested on a~single NVIDIA V100 card.} 

\begin{table}[h!]
\renewcommand\tabcolsep{5.7pt}
    \centering
    \begin{tabular}{lc}
     \toprule
         Hyperparameter & Value \\
         \midrule
         Optimiser & Adam \\
         & \cite{Kingma:2014} \\
         Learning rate & $0.002$\\
         $\beta_1$ and  $\beta_2$ & 0.9\\
         Number of epochs & 400 \\
         \hline
         BiLSTM layers & 2 \\
         BiLSTM dropout rate & 0.33 \\
         LSTM hidden size & 512 \\
         Arc projection size & 512 \\
         Label projection size & 128 \\
         \bottomrule
    \end{tabular}
    \caption{COMBO training parameters (the~upper entries) and model parameters (the~bottom entries).}
    \label{tab:hyper_train}
\end{table}

\section{Results}
The~shared task submissions are evaluated with two evaluation metrics: ELAS -- LAS\footnote{LAS (labelled attachment score) is the~proportion of tokens that are assigned the correct head and dependency label according to the gold standard.} on enhanced dependencies, and EULAS -- LAS on enhanced dependencies where labels are restricted to the~UD relation types, i.e. sublabels are ignored. COMBO ranks 4th, achieving 84.71\% ELAS in the~qualitative evaluation (an~average over treebanks), and 83.79\% ELAS in the~coarse evaluation (an~average over languages). In terms of EULAS, it ranks 4th achieving 86.30\% in the~qualitative evaluation, and 5th achieving 85.20\% in the~coarse evaluation. In addition to ELAS and EULAS metrics, the~systems are also compared in terms of quality of predicting labelled dependency trees measured with LAS (the~secondary evaluation measure). In the~LAS ranking, COMBO takes second place achieving 88.91\% in the~qualitative evaluation, and 87.84\% in the~coarse evaluation, being slightly overcome by the~ROBERTNLP system (89.25\% in the~qualitative evaluation, and 89.18\% in the~coarse evaluation). Table \ref{tab:results} presents the~official results of COMBO models per language.
\begin{table}[h!]
\renewcommand\tabcolsep{10pt}
\centering
\begin{tabular}{l|l|l|l}
\toprule
Language & LAS & EULAS & ELAS \\
\midrule
Arabic           & 81.04          & 78.35          & 76.39          \\
Bulgarian        & 89.52          & 87.41          & 86.67          \\
Czech            & 93.30          & 90.57          & 89.08          \\
Dutch            & 90.93          & 88.90          & 87.07          \\
English          & 87.22          & 85.27          & 84.09          \\
Estonian         & 87.53          & 85.56          & 84.02          \\
Finnish          & 92.28          & 88.79          & 87.28          \\
French           & 89.29          & 88.10          & 87.32          \\
Italian          & 93.27          & 91.16          & 90.40          \\
Latvian          & 90.25          & 86.22          & 84.57          \\
Lithuanian       & 84.75          & 81.28          & 79.75          \\
Polish           & 92.75          & 90.22          & 87.65          \\
Russian          & 94.29          & 91.76          & 90.73          \\
Slovak           & 91.72          & 88.53          & 87.04          \\
Swedish          & 87.82          & 85.26          & 83.20          \\
Tamil            & 56.28          & 53.49          & 52.27          \\
Ukrainian        & 90.96          & 87.60          & 86.92          \\
\hline
\textbf{Average} & \textbf{87.84} & \textbf{85.20} & \textbf{83.79} \\
\bottomrule
\end{tabular}
\caption{\label{tab:results} The official evaluation results per language.}
\end{table}

\begin{table}[h!]
\renewcommand\tabcolsep{4pt}
\centering
\begin{tabular}{l|l|l|l|l}
    \toprule
    \multirow{2}{*}{Language} & \multicolumn{2}{c}{Before} & \multicolumn{2}{c}{After} \\
    & EULAS & ELAS & EULAS & ELAS \\
    \midrule
    Arabic & 77.46 & 57.32 & 77.89 & 76.40 \\
    Bulgarian & 89.50 & 78.97 & 90.29 & 89.30 \\
    Czech & 89.93 & 74.96 & 91.28 & 89.91 \\
    Dutch & 87.96 & 76.22 & 88.94 & 87.64 \\
    English & 85.13 & 74.40 & 85.49 & 84.30 \\
    Estonian & 86.27 & 68.73 & 86.92 & 85.45 \\
    Finnish & 86.98 & 72.08 & 87.92 & 86.44 \\
    French & 90.48 & 89.99 & 91.10 & 90.62 \\
    Italian & 89.84 & 75.47 & 91.10 & 90.31 \\
    Latvian & 85.65 & 73.72 & 86.44 & 84.88 \\
    Lithuanian & 82.37 & 63.56 & 83.41 & 82.32 \\
    Polish & 90.08 & 77.97 & 90.64 & 87.64 \\
    Russian & 90.43 & 75.93 & 91.03 & 90.10 \\
    Slovak & 87.89 & 71.71 & 89.39 & 87.90 \\
    Swedish & 85.62 & 73.59 & 86.09 & 84.07 \\
    Tamil & 54.35 & 40.48 & 54.84 & 53.38 \\
    Ukrainian & 88.30 & 73.51 & 89.13 & 88.52 \\
    \bottomrule
\end{tabular}
\caption{\label{tab:postprocessing} Impact of the post-processing step.}
\end{table}

\paragraph{Post-processing impact}
We measure the impact of the post-processing step (i.e. extending graph labels with case information and correcting edges coming into the~function words) on the development data per language (see Table \ref{tab:postprocessing}). Following the~training approach, we concatenate the datasets if a language has multiple treebanks. 
The~second rule modifies the~graph structure. However, as the~EULAS scores are almost negligible, using this rule seems questionable. The~first rule, in turn, does not modify the~structure of EUD graphs, but only their edge labels, and its impact on improving ELAS scores is significant.

\begin{table}
\renewcommand\tabcolsep{6pt}
\centering
\begin{tabular}{l|l|l|l|l}
\toprule
\multirow{2}{*}{Language} & \multicolumn{2}{c}{Sentences} & \multicolumn{2}{c}{Tokens}\\ 
& TGIF & Stanza & TGIF & Stanza\\
\midrule
Arabic & 96.87 & 79.92 & 99.99 & 99.97\\ 
Dutch & 94.32 & 83.82 & 99.90 & 99.89 \\
Lithuanian & 96.22 & 87.74 & 99.99 & 99.81\\
Swedish & 99.03 & 93.64 & 99.86 & 99.44 \\
\bottomrule
\end{tabular}
\caption{\label{tab:segmentation_compar}
The~quality of TGIF and Stanza segmentation in the selected languages.}
\end{table}

\begin{table*}
\renewcommand\tabcolsep{10pt}
\centering
\begin{tabular}{l|l|l|l}
    \toprule
    Language & LAS & EULAS & ELAS \\
    \midrule
    Arabic           & 81.04 (+4.51)          & 78.35 (+4.3)          & 76.39 (+4.24)          \\
    Dutch            & 90.93 (+1.52)          & 88.90 (+1.61)          & 87.07 (+1.59)          \\
    Lithuanian       & 84.75 (+1.35)          & 81.28 (+1.34)          & 79.75 (+1.31)         \\
    Swedish          & 87.82 (+1.24)          & 85.26 (+1.21)          & 83.20 (+1.17)         \\
    \bottomrule
\end{tabular}
\caption{\label{tab:goldtoken}Performance gain in predicting UD trees and EUD graphs of gold-standard tokanised test sentences from the~languages with the worst segmentation quality. The~values in brackets show the improvement over the~baseline (i.e. Stanza tokenisation).}
\end{table*}

\paragraph{Segmentation drawback}
The~official evaluation results show significant discrepancies in the~quality of tokenisation and sentence segmentation. The~highest differences in sentence segmentation between TGIF, the~winner of the~shared task, and Stanza used in our approach 
are shown in Table \ref{tab:segmentation_compar}. For example, there is a~loss of more than 15 percentage points in sentence segmentation of the~Arabic texts. We therefore decide to investigate the~impact of the~quality of sentence segmentation and tokenisation on the~final results. For this purpose, we conduct an~additional experiment consisting in predicting EUD graphs on the~test data with gold-standard tokenisation and sentence segmentation. The~results of this experiment show a~gain of around 1.5 pp for all tested languages except Arabic with the~gain over 4 pp (see Table \ref{tab:goldtoken}).

\section{Conclusion}
We presented the~COMBO-based solution to EUD parsing which took part in the~IWPT 2021 EUD shared task. The~proposed approach is hybrid, i.e. based on machine learning and rule-based algorithms. 
First, UD trees and EUD graphs (and also morphosyntactic features of tokens, i.e. parts of speech, morphological features, and lemmata) are automatically predicted with the~data-driven COMBO system. Then, the~predicted structures are combined into the~EUD graphs using the~developed rule-based merge algorithm. Finally, the~labels of modifiers and conjuncts in the~merged EUD graphs are extended with case information using an~expansion rule. The~proposed solution is simple and language-independent. We recognise that we could still improve the~results, e.g. by defining language-specific correction rules. However, our objective was to build an~easy-to-use system for predicting EUD graphs that is publicly available and can be efficiently use to solve sophisticated NLU tasks.

\section*{Acknowledgments}
The research presented in this paper was founded by the~Polish Ministry of Education and Science as part of the investment in the CLARIN-PL research infrastructure. The~computing was performed at Pozna\'{n} Supercomputing and Networking Center.



\bibliographystyle{acl_natbib}
\bibliography{acl2021,anthology}


\end{document}